\PassOptionsToPackage{pdftex}{graphicx} 
\documentclass{article}

\PassOptionsToPackage{numbers, compress}{natbib}
\usepackage[preprint]{neurips_2023}

\usepackage[utf8]{inputenc} 
\usepackage[T1]{fontenc}    
\usepackage{hyperref}       
\usepackage{url}            
\usepackage{booktabs}       
\usepackage{amsfonts}       
\usepackage{nicefrac}       
\usepackage{microtype}      
\usepackage{xcolor}         
\usepackage{adjustbox}
\usepackage{float}
\usepackage{svg}

\usepackage{tikz}
\usetikzlibrary{quotes,positioning}
\tikzset{
    simple node image/.style n args={0}{%
        rectangle,
        inner sep=0,
        text centered,
        anchor=center,
        align=center,
        node distance=0mm
    }
}

\usepackage{xcolor}
\usepackage{listings}
\usepackage{siunitx}
\usepackage{MnSymbol}
\newcommand{\bv}{\mathbf{v}}
\usepackage{cleveref}

\title{Improvements to SDXL in NovelAI Diffusion V3}

\author{%
  Juan Ossa \\
  Anlatan LLC \\
  \texttt{juan@anlatan.ai} \\
  \And
  Eren Doğan \\
  Anlatan LLC \\
  \texttt{eren@anlatan.ai} \\
  \And
  Alex Birch \\
  Anlatan LLC \\
  \And
  F. Johnson\\
  Anlatan LLC
}

\begin{document}

\maketitle

\begin{abstract}
  In this technical report, we document the changes we made to SDXL in the process of training NovelAI Diffusion V3, our state of the art anime image generation model.
\end{abstract}

\section{Introduction}
Diffusion based image generation models have been soaring in popularity recently, with a variety of different model architectures being explored. One such model, Stable Diffusion, has achieved high popularity after being released as Open Source. Following up on it, Stability AI released SDXL, a larger and extended version following its general architecture \cite{podell2023sdxl}. We chose SDXL as the basis for our latest image model, NovelAI Diffusion V3, and made several enhancements to its training practices.

This technical report is structured as follows. In Section~\ref{sec:enhancements}, we describe our enhancements in detail. Following that, we evaluate our contributions in Section~\ref{sec:results}. Finally, we draw conclusions in Section~\ref{sec:conclusions}.

\section{Enhancements}
\label{sec:enhancements}
In this section, we present the enhancements we applied to SDXL to improve generation results.

\subsection{v-Prediction Parameterization}
\label{sec:vpred}
We uptrained SDXL from $\epsilon$-prediction to $\bv$-prediction\cite{salimans2022progressivedistillationfastsampling} parameterization. This was instrumental to our goal of supporting Zero Terminal SNR (see \cref{sec:terminalnoise}). The $\epsilon$-prediction objective ("where's the noise?") is trivial at SNR=0 ("everything is noise"), hence $\epsilon$ loss fails to teach the model how to predict an image from pure noise. By comparison, $\bv$-prediction transitions from $\epsilon$-prediction to $x_0$-prediction as appropriate, ensuring that neither high nor low SNR timesteps are trivially predicted.

As a secondary benefit, we sought to access the merits described in \cite{ho2022imagenvideohighdefinition}: improved numerical stability, elimination of colour-shifting\cite{yu2024unmaskingbiasdiffusionmodel} at high resolutions, and faster convergence of sample quality.

\subsection{Zero Terminal SNR}
\label{sec:terminalnoise}
Stable-diffusion\cite{rombach2022highresolutionimagesynthesislatent} and SDXL were trained with a flawed noise schedule\cite{lin2024commondiffusionnoiseschedules}, limiting image-generation to always produce samples with medium brightness. We aimed to remedy this.

\begin{figure}[H]
    \centering
    \adjustbox{max width=\columnwidth}{
    \input{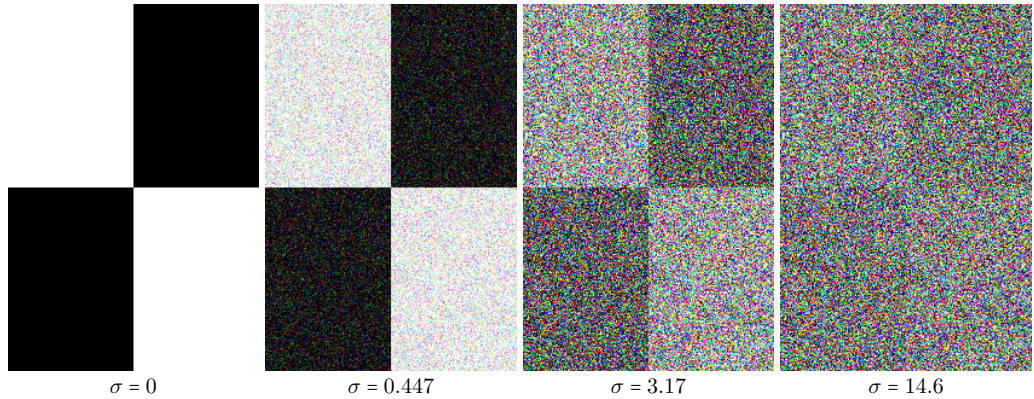}
    }
    \caption{Noise is added to a sample, until the final training timestep $\sigma_\text{max}$, where Gaussian noise with standard deviation 14.6 is added. This amount of noise does not sufficiently destroy the signal in the image; the lowest frequencies (in particular its average colour) remain discernable.}
    \label{fig:noising_visual}
\end{figure}

\begin{figure}[H]
    \centering
    \adjustbox{max width=\columnwidth}{
    \input{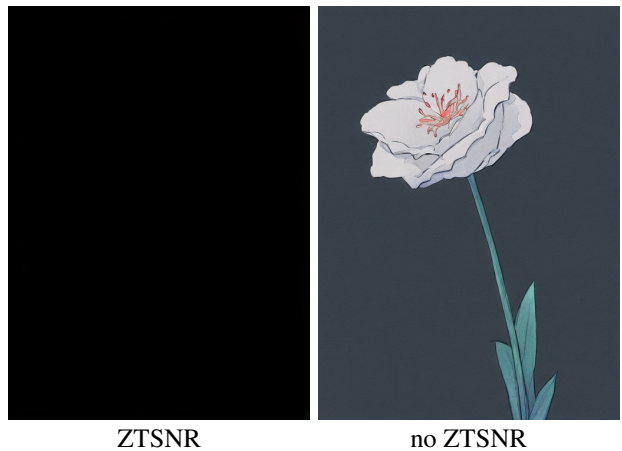}
    }
    \caption{Prompting the model to generate "completely black". A model trained to predict images from infinite-noise (ZTSNR) can comply with the prompt. Whereas if we begin inference from a timestep with finite noise, the model outputs an image with medium brightness, trying to match the mean colour it sees in the starting noise, and consequently generates a non-prompt-relevant sample.}
    \label{fig:black_visual}
\end{figure}

Diffusion models\cite{ho2020denoisingdiffusionprobabilisticmodels} such as these learn to reverse an information-destroying process (typically\cite{bansal2022colddiffusioninvertingarbitrary} the application of Gaussian noise). A principled implementation of diffusion should employ a noise schedule that spans from pure-signal to pure-noise. Unfortunately SDXL's noise schedule stops short of pure-noise (\cref{fig:noising_visual}). This teaches the model a bad lesson: "there is always some signal in the noise". This assumption is harmful at inference-time, where our starting point is pure-noise, and results in the creation of non-prompt-relevant features (\cref{fig:black_visual}).

\begin{figure}[H]
    \centering
    \adjustbox{max width=\columnwidth}{
    \input{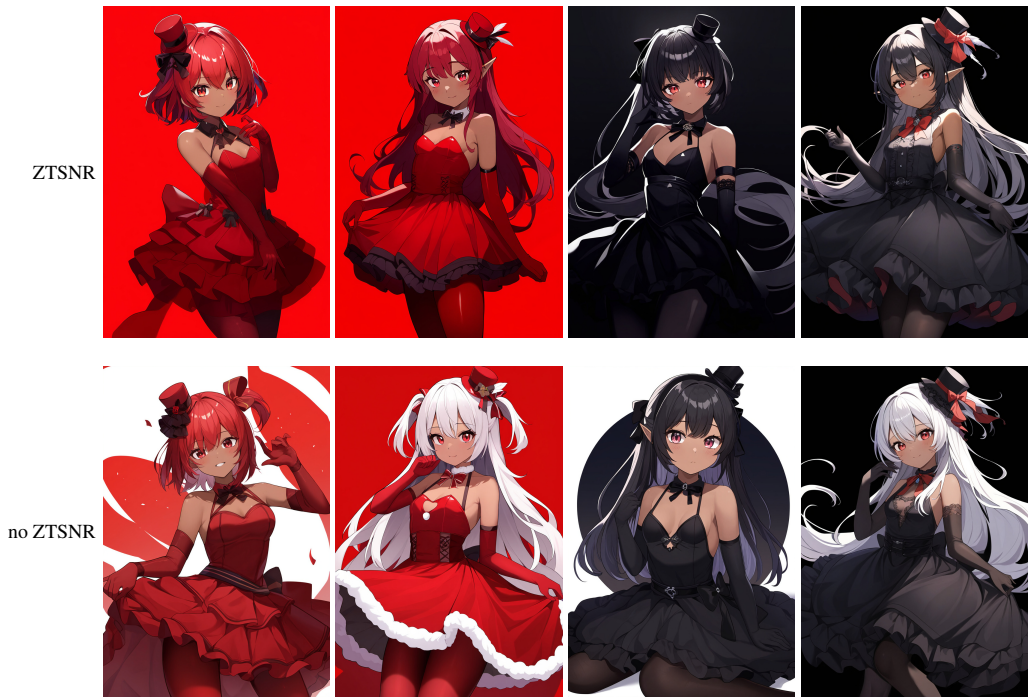}
    }
    \caption{Absence of a ZTSNR step introduces spurious high-contrast coarse features, attempting to pull the canvas's mean (latent) colour back to 0, the average value of the Gaussian noise provided at the start of inference. Concretely, this can mean an opposing colour is added to the background, or hair colour and clothing details can disobey the prompt.}
    \label{fig:ztsnr_visual}
\end{figure}

We trained NAIv3 on a noise schedule with Zero Terminal SNR, to expose SDXL to pure-noise during training. We train the model up to noise levels so high that it can no longer rely on mean-leakage, and learns to predict relevant mean colours and low frequencies from the text condition (\cref{fig:ztsnr_visual}) instead. The introduction of a ramp up to infinite-noise aligns the training schedule with the inference schedule.

\begin{figure}[H]
    \centering
    \adjustbox{max width=\columnwidth}{
    \input{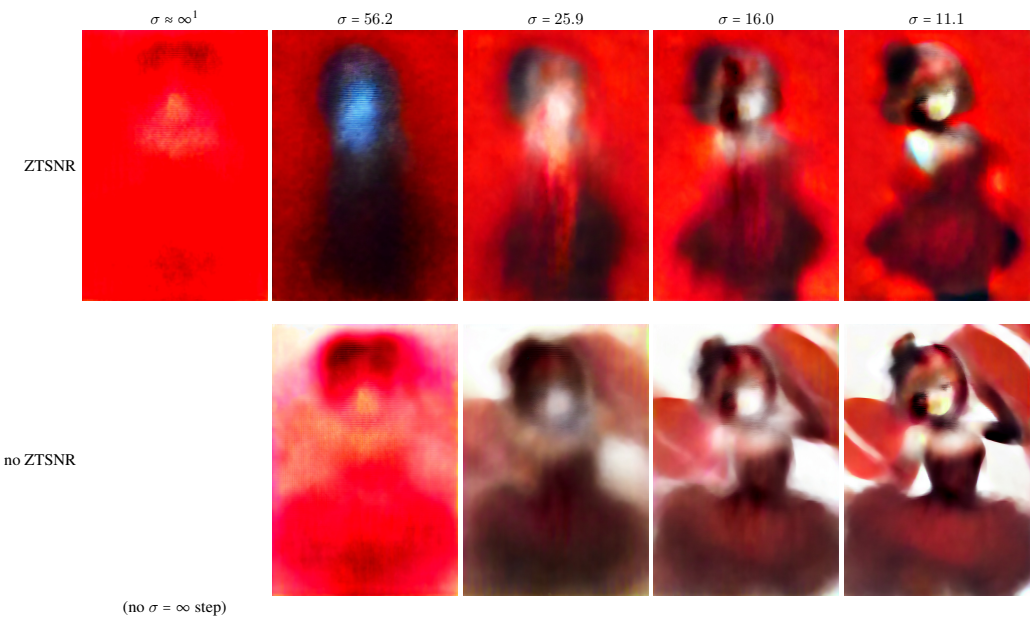}
    }
    \caption{Intermediate denoising predictions with and without ZTSNR. The ZTSNR regime proposes a relevant average canvas colour at $\sigma\approx\infty$\footref{fnote:terminal-sigma}, whereas the non-ZTSNR regime understands that in a signal noised up only to $\sigma=56$, the mean colour should still be discernable, and incorrectly concludes that the average colour should be 0, adding white to the canvas in order to achieve this.}
    \label{fig:ztsnr_intermediates_visual}
\end{figure}

ZTSNR poses practicality issues when implemented in the EDM\cite{karras2022elucidatingdesignspacediffusionbased} framework (i.e. via the k-diffusion\cite{katherine_crowson_2023_10284390} library). We explain in \cref{sec:ztsnr-practicalities} how to overcome such issues.

\subsection{Sampling from High-Noise Timesteps Improves High-Resolution Generation}
\begin{figure}[H]
    \centering
    \adjustbox{max width=\columnwidth}{
    \includegraphics{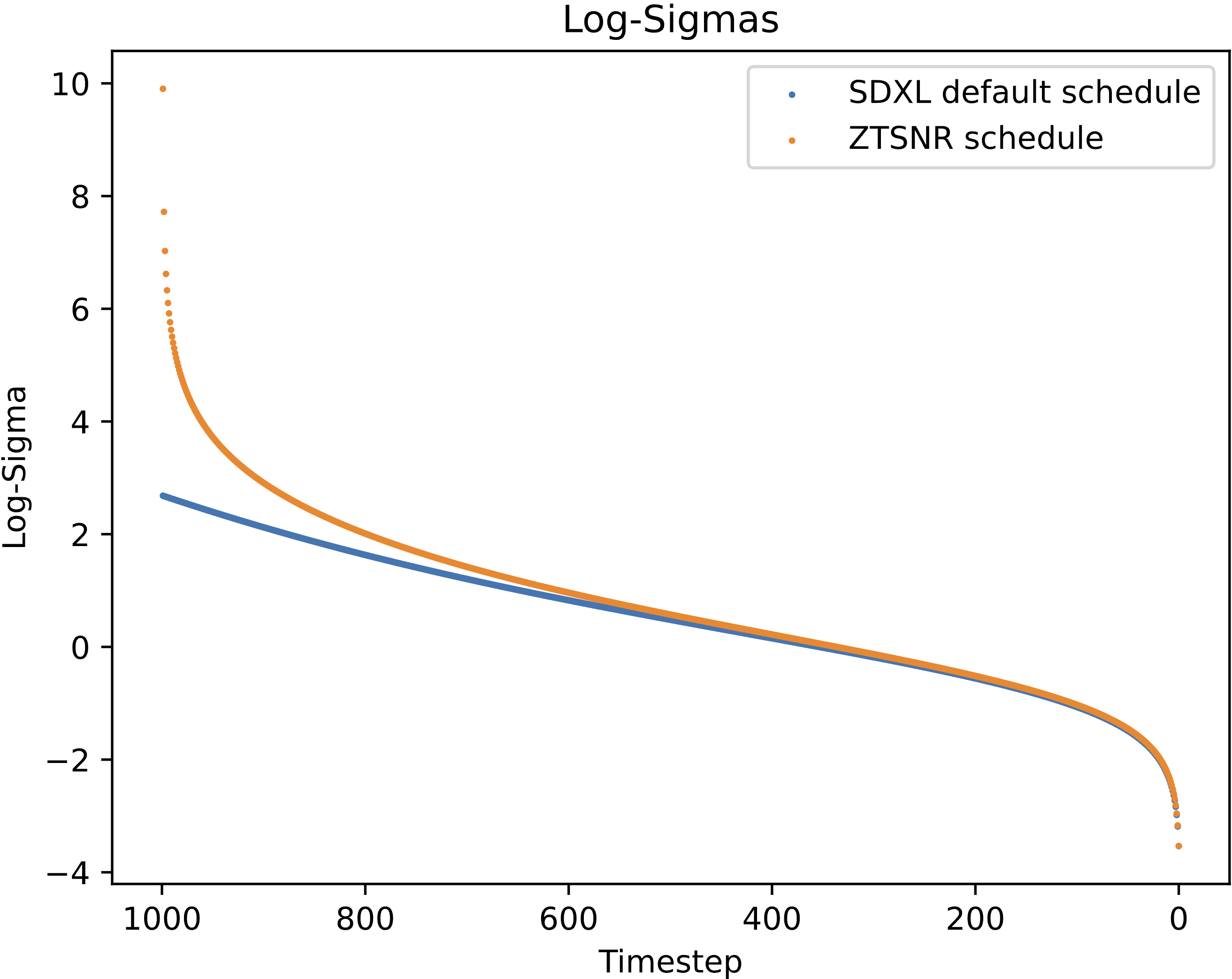}
    }
    \caption{Schedule with/without ZTSNR.}
    \label{fig:native_schedule_comparison}
\end{figure}

Curving the schedule to converge on $\sigma\approx\infty$\footref{fnote:terminal-sigma} confers a ramp of higher sigmas along the way (\cref{fig:native_schedule_comparison}). This helps to resolve another problem with SDXL: its $\sigma_\text{max}$ is not high enough to destroy the low frequencies of signal in high-resolution images. In fact $\sigma_\text{max}$ was not increased since SD1\cite{rombach2022highresolutionimagesynthesislatent}, despite the increase in target resolution. As resolution increases (or rather as the amount of redundant signal increases — increasing latent channels could be another way to increase redundancy): more noise is required to achieve a comparable SNR (i.e. destroy an equivalent proportion of signal)\cite{hoogeboom2023simplediffusionendtoenddiffusion}.

\begin{figure}[H]
    \centering
    \adjustbox{max width=\columnwidth}{
    \input{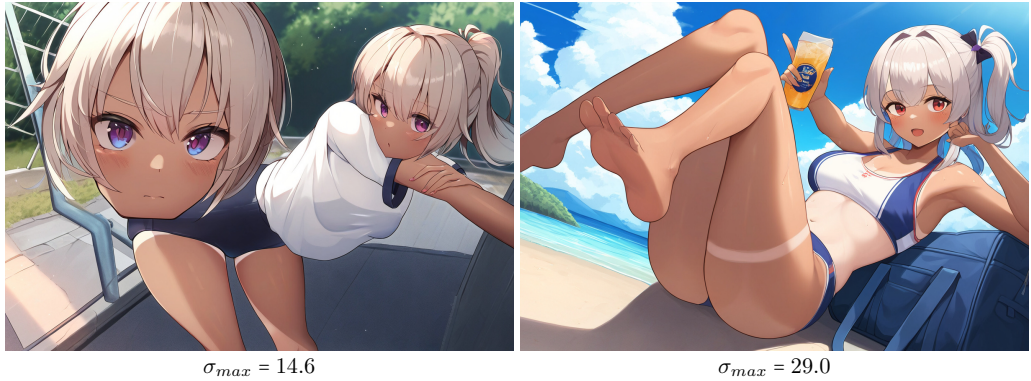}
    }
    \caption{Effect of raising $\sigma_\text{max}$. SDXL's default $\sigma_\text{max}=14.6$ is insufficient for global coherence on high-resolution images, resulting in multi-body artifacting. Doubling to $\sigma_\text{max}=29.0$ resolves this artifacting. In practice we raise $\sigma_\text{max}$ even higher than this, to $\sigma_\text{max}\approx\infty$\footref{fnote:terminal-sigma}, to achieve a relevant mean colour too.}
    \label{fig:sigma_max_visual}
\end{figure}

Large features, such as close-to-camera limbs can lose coherence in SDXL's default regime of $\sigma_\text{max}=14.6$. Doubling $\sigma_\text{max}$ (for example) recovers coherence (\cref{fig:sigma_max_visual}). In practice we hit higher sigmas than this in our standard 28-step native schedule, which is a uniform linear spacing of timesteps over the 1000-step ZTSNR schedule, the highest of which is approx. $\sigma_\infty$
\footnote{\label{fnote:terminal-sigma}For the $\sigma=\infty$ step, in practice we use an approximation $\sigma=20000\approx\infty$. See \cref{sec:ztsnr-practicalities} for the rationale behind this practical approximation, and a more principled approach.}.

\begin{figure}[H]
    \centering
    \adjustbox{max width=\columnwidth}{
    \input{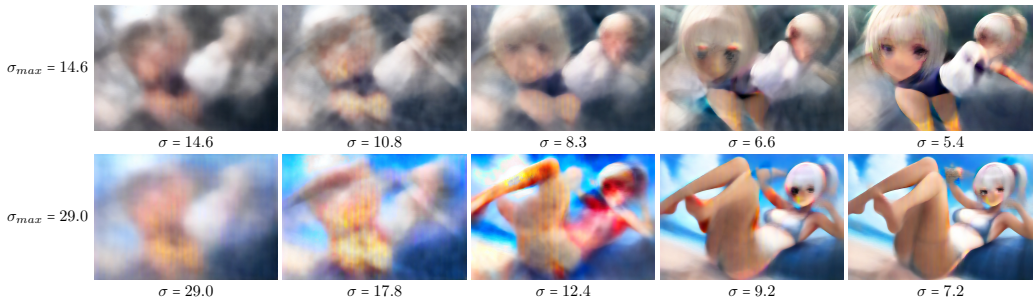}
    }
    \caption{Intermediate denoising predictions, on different $\sigma_\text{max}$ regimes.}
    \label{fig:sigma_max_intermediates_visual}
\end{figure}

A rule of thumb is that if you double the canvas length (quadrupling the canvas area): you should double $\sigma_\text{max}$ (quadrupling the noise variance) to maintain SNR. This is an upper bound which assumes that the extra signal is fully redundant. The approximation is better for high-resolution images, and sufficient for our purposes.

More recently, a non-commercial, research-licensed version of SDXL (CosXL)\cite{cosxl} has been released with v-prediction and ZTSNR support.

\subsection{MinSNR}
\label{sec:minsnr}
We used MinSNR\cite{hang2024efficientdiffusiontrainingminsnr} loss-weighting to treat diffusion as a multi-task learning problem, balancing the learning of each timestep according to difficulty, and avoiding focusing too much training on low-noise timesteps.

\section{Dataset}
\label{sec:dataset}
Our dataset consisted of approximately 6 million images gathered from crowd-sourced platforms. It was enriched with highly detailed, tag-based labels. The images are mostly illustrations in styles commonly found in Japanese animation, games and pop-culture.

\section{Training}
\label{sec:training}
We trained the model on our 256x H100 cluster for many epochs and roughly 75k H100 hours. We also used a staged approach, with later stages consisting of more curated, higher quality data. We trained in float32, with tf32\cite{tensorfloat} optimization. Our total compute budget was above that of the original SDXL training run, allowing us to thoroughly adapt the model to our data's distribution.

Adaptation to the changes described in Section~\ref{sec:enhancements} was quite fast. While the initial step, initialized from the original SDXL weights, showed only noise, our first set of samples, output after about 30 minutes (wall time) of training, was already coherent.

\subsection{Aspect-Ratio Bucketing}
As in previous NovelAI Diffusion models\cite{anlatanimprovestablediffusion}, we prepared like-aspect minibatches via aspect-ratio bucketing. This enabled us to frame images better than in a center-crop regime, and achieve better token-efficiency than in a padding regime. In \cref{sec:arbreasoning,sec:arbbatches,sec:arbloading}, we will reiterate our approach here for the sake of completeness.

\subsubsection{Reasoning}
\label{sec:arbreasoning}
Existing image generation models\cite{rombach2022highresolutionimagesynthesislatent} are very prone to producing images with unnatural crops. This is due to training practicality: uniform batches are simple to implement. Often practitioners opt to train on square data, taking crops from the center of the image. This is not conducive to modeling typical image data distributions, as most photos and artworks are not square.

As a consequence, humans are often generated without feet or heads, and swords consist of only a blade with a hilt and point outside the frame. As we are creating an image generation model to accompany our storytelling experience, it is important for our model to be able to produce proper, uncropped characters, and generated knights should not be holding a metallic-looking straight line extending to infinity.

Another issue with training on cropped images is that it can lead to a mismatch between the text and the image.

\begin{figure}[H]
    \centering
    \adjustbox{max height=100mm}{
    \includegraphics{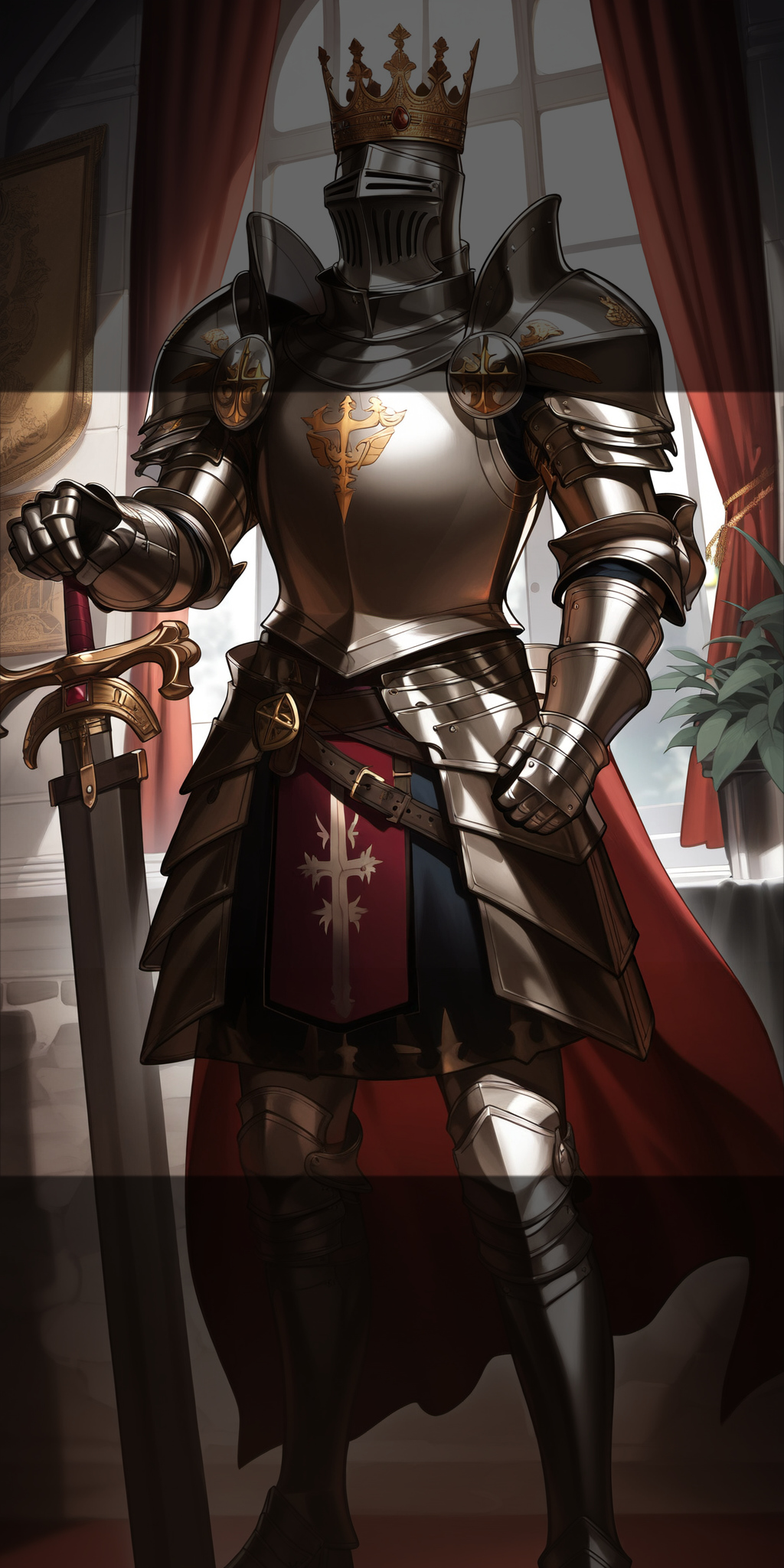}
    }
    \caption{Center-cropping (demonstrated here with dark letterboxing) can remove details crucial for prompt-relevance; in this case "crown" has been excluded from the image.}
    \label{fig:center_crop}
\end{figure}

For example, an image with a "crown" tag will often no longer contain a crown after a center crop is applied, leaving the monarch thereby decapitated (\cref{fig:center_crop}).

We found that using random crops instead of center crops only slightly improves these issues.

Using Stable Diffusion with variable image sizes is possible, although it can be noticed that going too far beyond the native resolution of $512\times512$ tends to introduce repeated image elements, and very low resolutions produce indiscernible images. This is likely to be due to its lack of an explicit position embedding, resulting in a model which determines position from side-channels like convolution padding\cite{islam2021positionpaddingpredictionsdeeper,karras2021aliasfreegenerativeadversarialnetworks}. Some training-free approaches seek to generalize inference to resolutions outside of the training distribution via techniques such as convolution dilation \cite{huang2024fouriscalefrequencyperspectivetrainingfree,he2023scalecraftertuningfreehigherresolutionvisual,wu2024megafusionextenddiffusionmodels} and attention entropy scaling \cite{jin2023trainingfreediffusionmodeladaptation,chiang2022overcomingtheoreticallimitationselfattention}. We chose instead to train on varied image sizes.

Training on single, variable sized samples would be trivial, but also extremely slow and more liable to training instability due to the lack of regularization provided by the use of mini batches.

\subsubsection{Custom Batch Generation}
\label{sec:arbbatches}
As no existing solution for this problem seems to exist, we have implemented custom batch generation code for our dataset that allows the creation of batches where every item in the batch has the same size, but the image size of batches may differ.

We do this through a method we call aspect ratio bucketing. An alternative approach would be to use a fixed image size, scale each image to fit within this fixed size and apply padding that is masked out during training. Since this leads to unnecessary computation during training, we have not chosen to follow this alternative approach.

In the following, we describe the original idea behind our custom batch generation scheme for aspect ratio bucketing.

First, we have to define which buckets we want to sort the images of our dataset into. For this purpose, we define a maximum image size of $512\times768$ with a maximum dimension size of 1024. Since the maximum image size is $512\times768$, which is larger than $512\times512$ and requires more VRAM, per-GPU batch size has to be lowered, which can be compensated through gradient accumulation.

We generate buckets by applying the following algorithm:

\begin{itemize}
    \item Set the width to 256.
    \item While the width is less than or equal to 1024:
    \begin{itemize}
     \item Find the largest height such that height is less than or equal to 1024 and that width multiplied by height is less than or equal to $512 \cdot 768$.
     \item Add the resolution given by height and width as a bucket.
     \item Increase the width by 64.
    \end{itemize}
\end{itemize}

The same is repeated with width and height exchanged. Duplicated buckets are pruned from the list, and an additional bucket sized $512\times512$ is added.

Next, we assign images to their corresponding buckets. For this purpose, we first store the bucket resolutions in a NumPy array and calculate the aspect ratio of each resolution. For each image in the dataset, we then retrieve its resolution and calculate the aspect ratio. We compute the logarithms of each aspect ratio, in order to compare in log-space\footnote{aspect ratios compared in log-space have the desirable property that a 1:1 aspect ratio can be equidistant from buckets 2:1 and 1:2, each of which fit an equal proportion of its area.}. The image angle is subtracted from the array of bucket angles, allowing us to efficiently select the closest bucket according to the absolute value of the difference between aspect ratios.

\begin{equation}\label{eq:arb}
    \text{image\_bucket} = \text{argmin}(\text{abs}(\text{log}(\text{bucket\_aspects}) - \text{log}(\text{image\_aspect}))))
\end{equation}

The image’s bucket number is stored associated with its item ID in the dataset. If the image’s aspect ratio is very extreme and too different from even the best-fitting bucket, the image is pruned from the dataset.

Since we train on multiple GPUs, before each epoch, we shard the dataset to ensure that each GPU works on a distinct subset of equal size. To do this, we first copy the list of item IDs in the dataset and shuffle them. If this copied list is not divisible by the number of GPUs multiplied by the batch size, the list is trimmed, and the last items are dropped to make it divisible.

We then select a distinct subset of $\frac{1}{\text{world\_size}*\text{bsz}}$ item IDs according to the global rank of the current process. The rest of the custom batch generation will be described as seen from a single shard of these processes and operate on the subset of dataset item IDs.

For the current shard, lists for each bucket are created by iterating over the list of shuffled dataset item IDs and assigning each ID to the list corresponding to the bucket that best fits that image's aspect ratio.

Once all images are processed, we iterate over the lists for each bucket. If its length is not divisible by the batch size, we remove the last elements on the list as necessary to make it divisible and add them to a separate catch-all bucket. As the overall shard size is guaranteed to contain a number of elements divisible by the batch size, this process is guaranteed to produce a catch-all bucket with a length divisible by the batch size as well.

When a batch is requested, we randomly draw a bucket from a weighted distribution. The bucket weights are set as the size of the bucket divided by the size of all remaining buckets. This ensures that even with buckets of widely varying sizes, the custom batch generation does not introduce bias during training. If buckets were chosen without weighting, small buckets would empty out early during the training process, and only the biggest buckets would remain towards the end of training.

A batch of items is finally taken from the chosen bucket. The items taken are removed from the bucket. If the bucket is now empty, it is deleted for the rest of the epoch. The chosen item IDs and the chosen bucket’s resolution are now passed to an image-loading function.

\subsubsection{Image Loading}
\label{sec:arbloading}
Each item ID’s image is loaded and processed to fit within the bucket resolution. For fitting the image, two approaches are possible.

First, the image could be simply rescaled. This would lead to a slight distortion of the image. For this reason, we have opted for the second approach:

The image is scaled, while preserving its aspect ratio, in such a way that it:

\begin{itemize}
    \item Either fits the bucket resolution exactly if the aspect ratio happens to match,
    \item or it extends past the bucket resolution on one dimension while fitting it exactly on the other.
\end{itemize}
In the latter case, a random crop is applied.

As we found that the mean aspect ratio error per image is only 0.033, these random crops only remove very little of the actual image, usually less than 32 pixels.

The loaded and processed images are finally returned.

\subsection{Conditioning}
Like in our previous models, we used CLIP\cite{radford2021learningtransferablevisualmodels} context concatenation and conditioned on CLIP's penultimate hidden states (on this part, no change was required; SDXL base already does this). To produce SDXL's pooled CLIP condition given multiple concatenated CLIP contexts: we take a mean average over all CLIP segments' pooled states.

\subsection{Tag-based Loss Weighting}
\label{sec:tlm}
During training, we employed a tag-based loss weighting scheme, which keeps track of how often tags in certain tag type classes occur. Images with tags that are overly common within their class have their loss downweighted, while images with tags that are rare within their class may have their loss up-weighted. This allows our model to better learn concepts from rare tags while reducing the influence of over-represented concepts.

\subsection{VAE Decoder Finetuning}
As in NovelAI Diffusion V1, we finetune the Stable-Diffusion (this time SDXL) VAE decoder, which decodes the low-resolution latent output of the diffusion model, into high-resolution RGB images. The original rationale (in V1 era) was to specialize the decoder for producing anime textures, especially eyes. For V3, an additional rationale emerged: to dissuade the decoder from outputting spurious JPEG artifacts, which were being exhibited despite not being present in our input images.

\section{Results}
\label{sec:results}
We find empirically that our model produces relevant, coherent images at CFG\cite{ho2022classifierfreediffusionguidance} scales between 3.5–5. This is lower than the default of 7.5 recommended typically for SDXL inference, and suggests that our dataset is better-labelled.

\section{Conclusions}
\label{sec:conclusions}
NovelAI Diffusion V3 is our most successful image generation model yet, generating 4.8M images per day. From this strong base model we have been able to uptrain a suite of further products, such as Furry Diffusion V3, Director Tools, and Inpainting models.

\section{Acknowledgements}
Thanks to Stefan Baumann for the ZTSNR derivations and for reviewing this report. Thanks also to Tanishq Abraham for review. Additionally, we would like to thank Stefan Baumann and Katherine Crowson for explaining practical approaches for implementing ZTSNR in k-diffusion\cite{katherine_crowson_2023_10284390}, and for explaining heuristics to scale $\sigma_\text{max}$ with image resolution.

\bibliographystyle{IEEEtranS}
\bibliography{bibliography}

\newpage
\appendix
\clearpage
\section{Practicalities for Implementing ZTSNR in k-diffusion/EDM}
\label{sec:ztsnr-practicalities}

ZTSNR poses practicality issues when implemented in the EDM\cite{karras2022elucidatingdesignspacediffusionbased} framework (i.e. via the k-diffusion\cite{katherine_crowson_2023_10284390} library). Models using the EDM formulation are expected to accept VE (Variance-Exploding, in this case $\sigma=\infty$) noise, but this poses numeric representability issues. Likewise, EDM-space samplers are forced to take an infinitely large Euler step with an infinitesimally small velocity. We discuss two solutions for how overcome such practical issues.

\subsection{Principled Implementation of ZTSNR}

The usual prescription of k-diffusion/EDM is to start with VE noise and scale it to VP (Variance-Preserving, i.e. unit variance). Practically, we can neither numerically represent a $\sigma=\infty$ random tensor nor can we divide it by $\infty$ to convert it to a variance-preserving formulation. We must find a way to skip this step. The only problems are practical/numeric; algebraically we possess alternative ways to compute this.

First we will need a bypass within the k-diffusion model wrapper, to "pass unit-variance pure-noise directly to the model", instead of converting $\sigma=\infty$ VE noise to VP noise. We also need to add a special case to discretization, to map $\sigma=\infty$ to the index of the maximum timestep.

Next we will need to special-case how we sample from $\sigma=\infty$; this is the first step of a ZTSNR inference schedule (though we may skip it if we are doing img2img\cite{meng2022sdeditguidedimagesynthesis}).

\subsubsection{Sampling from the Infinite-Noise Timestep}

During inference, we sample with the goal of predicting the clean image. This is not the objective our neural network was trained on. SDXL was trained to predict the noise present in the sample. And in NAIv3 we adapted SDXL (\cref{sec:vpred}) to instead predict velocity $\mathbf{v}$\cite{salimans2022progressivedistillationfastsampling}, an objective which transitions from noise-prediction to image-prediction as SNR changes.

k-diffusion provides model wrappers to adapt our neural network $F_\theta(\cdot)$ (which may predict noise, velocity, or some other objective) into an abstract denoiser $D_\theta$ which predicts the clean image. Concretely, the wrapper invokes the model through the parameterization of the Karras preconditioner\cite{karras2022elucidatingdesignspacediffusionbased}. On top of this $D_\theta$ denoiser abstraction, k-diffusion provides samplers which iterate on the clean-image predictions, converging on a fully-denoised image.

The Karras preconditioner describes how a clean-image denoiser $D_\theta$ can be related to our trained network $F_\theta(\cdot)$ through noise level $\sigma$-dependent scaling functions $c_{\mathrm{skip}}(\sigma)$, $c_{\mathrm{out}}(\sigma)$, and $c_{\mathrm{in}}(\sigma)$.
\begin{equation}
    D_{\theta}(\mathbf{x};\sigma)= c_{\mathrm{skip}}(\sigma)\mathbf{x}+c_{\mathrm{out}}(\sigma)F_{\theta}(c_{\mathrm{in}}(\sigma)\mathbf{x};\sigma),\label{eq:preconditioner}
\end{equation}

These scaling functions allow our model $F_\theta(\cdot)$ to transition between predicting noise content and clean sample as the noise level changes. By assigning values to these scaling functions, we can expose which terms can be eliminated at $\sigma=\infty$.

To sample an image from $D_\theta$, we can take Euler steps to bring the noise level down from $\sigma_i$ to $\sigma_{i + 1}$, $\sigma_{i+1} < \sigma_i$,
\begin{equation}\label{eq:euler_step}
    \mathbf{x}_{i+1} \leftarrow \mathbf{x}_i + (\sigma_{i + 1}-\sigma_{i})\cdot\frac{\mathbf{x}_i-D_{\theta}(\mathbf{x}_i;\sigma_i)}{\sigma_i}.
\end{equation}

and in so doing iterate towards a clean image. But in its current form, we cannot handle a Euler step down from $\sigma=\infty$. The step size $\sigma_{i + 1}-\sigma_{i_\infty}$ would be infinite, and the derivative $\frac{\mathbf{x}_i-D_\theta(\cdot)}{\sigma_\infty}$ has infinite terms on its numerator and denominator. The infinite numerator is due to the noised image $\mathbf{x}_i$ being formulated as VE, which at $\sigma=\infty$ entails infinite variance. Moreover the step is accumulated by summation with an infinite term $\mathbf{x}_i$.

We can make this tractable by formulating a special-case Euler step for the first step $i = 0$ of sampling. We decompose the initial VE noise into a standard Gaussian and its standard deviation, via the relationship $\mathbf{x}_0 = \sigma_0\mathbf{n}$, where $\mathbf{n} \sim \mathcal{N}(0, \mathbf{I})$. By assigning infinite standard deviation $\sigma_0 \rightarrow \infty$, we can eliminate $\sigma_0$ terms outside the denoiser $D_\theta$ by rearranging \cref{eq:euler_step} as follows:
\begin{align}
    \mathbf{x}_{1} &\leftarrow \sigma_0\mathbf{n} + (\sigma_1-\sigma_0)\cdot\frac{\sigma_0\mathbf{n}-D_{\theta}(\sigma_0\mathbf{n};\sigma_0)}{\sigma_0}\\
    &= \sigma_0\mathbf{n} + (\sigma_1-\sigma_0)\cdot\left(\mathbf{n} - \frac{D_{\theta}(\sigma_0\mathbf{n};\sigma_0)}{\sigma_0}\right)\\
    &= \sigma_0\mathbf{n} + \left(\sigma_1\mathbf{n} - \sigma_0\mathbf{n} - (\sigma_1-\sigma_0)\cdot\frac{D_{\theta}(\sigma_0\mathbf{n};\sigma_0)}{\sigma_0}\right)\\
    &= \sigma_1\mathbf{n} - (\sigma_1-\sigma_0)\cdot\frac{D_{\theta}(\sigma_0\mathbf{n};\sigma_0)}{\sigma_0}\\
    &= \sigma_1\mathbf{n} - \frac{\sigma_1-\sigma_0}{\sigma_0}\cdot D_{\theta}(\sigma_0\mathbf{n};\sigma_0)\\
    &= \sigma_1\mathbf{n} - \frac{\sigma_1}{\sigma_0}\cdot D_{\theta}(\sigma_0\mathbf{n};\sigma_0) + D_{\theta}(\sigma_0\mathbf{n};\sigma_0)\\
    &= \sigma_1\mathbf{n} + D_{\theta}(\sigma_0\mathbf{n};\sigma_0)\label{eq:euler_simplified_pt1_last}
\end{align}
Next we expand the Karras preconditioner $D_\theta$ to reveal its relationship with the underlying neural network $F_\theta$. Posing the Euler step in terms of these scaling factors will expose $\sigma_0$ terms, and create an opportunity to eliminate them as $\sigma_0 \rightarrow \infty$. In our case, $F_\theta$ has a v-prediction objective. The Karras preconditioner can be adapted to a v-prediction model using very similar scalings to the EDM denoiser, except with a negative $c_{\mathrm{out}}(\sigma)$ \cite{katherine_crowson_2023_10284390}:
\begin{align}
    c_\mathrm{skip}(\sigma) &= \frac{\sigma_\mathrm{data}^2}{\sigma^2 + \sigma_\mathrm{data}^2}\\
    c_{\mathrm{out}}(\sigma) &= \frac{-\sigma \cdot \sigma_{\mathrm{data}}}{\sqrt{\sigma_{\mathrm{data}}^2 + \sigma^2}}\\
    c_{\mathrm{in}}(\sigma) &= \frac{1}{\sqrt{\sigma^2 + \sigma_{\mathrm{data}}^2}}\label{eq:scalings}
\end{align}
These scaling factors simplify as $\sigma_0 \rightarrow \infty$, giving us
\begin{align}
    c_\mathrm{skip}(\sigma) &= 0\\
    c_{\mathrm{out}}(\sigma) &= -\sigma_\mathrm{data},
\end{align}
we can pose our Euler step \cref{eq:euler_simplified_pt1_last} in terms of the Karras preconditioner \cref{eq:preconditioner}, then apply these scalings (\cref{eq:scalings}) to eliminate all $\sigma_0$ terms outside the neural network $F_\theta$:
\begin{align}
    \mathbf{x}_{1} &\leftarrow \sigma_1\mathbf{n} + c_{\mathrm{skip}}(\sigma_0)\sigma_0\mathbf{n}+c_{\mathrm{out}}(\sigma_0)F_{\theta}(c_{\mathrm{in}}(\sigma_0)\sigma_0\mathbf{n}; \sigma_0)\\
    &= \sigma_1\mathbf{n} - \sigma_\mathrm{data}F_{\theta}(c_{\mathrm{in}}(\sigma_0)\sigma_0\mathbf{n}; \sigma_0)\\
    &= \sigma_1\mathbf{n} - \sigma_\mathrm{data}F_{\theta}\left(\frac{\sigma_0}{\sqrt{\sigma_0^2 + \sigma_{\mathrm{data}}^2}}\mathbf{n}; \sigma_0\right)\\
    &= \sigma_1\mathbf{n} - \sigma_\mathrm{data}F_{\theta}(\mathbf{n}; \sigma_0).
\end{align}
Now, by special-casing the noise level conditioning mechanism for the zero-terminal SNR step $\sigma_0$, the first sampling step no longer contains any non-finite terms and thus can directly be implemented.

After completing this ZTSNR step, we would hand over to a conventional sampler (in much the same way as img2img\cite{meng2022sdeditguidedimagesynthesis} is implemented) to continue denoising from a finite sigma. We do not attempt to incorporate this ZTSNR step into a multistep (such as in DPM++ 2M \cite{lu2023dpmsolverfastsolverguided}), as this would expose us to the "infinite step size" problem again. This is not a big loss; even if we were to implement a ZTSNR multistep, the step size ratio $\frac{h_\mathrm{last}}{h}$ would be infinite, so a DPM++ multistep (which incorporates former steps in inverse proportion to this step size ratio) would use an infinitessimal amount of the ZTSNR step's estimate.

\subsection{Trivial Implementation of (Almost) Zero Terminal SNR}

We opted for a simpler solution: use k-diffusion as usual, but clamp the sigmas by which our schedule is defined, to a maximum of 20000. Noise is drawn in float32, as is the scaling of VE noised latents to VP. $\sigma=20000$ is more than sufficient to destroy the lowest frequencies of the noise. Empirically, we find that even a much lower sigma of somewhere around 136–317 is sufficient to prevent our model expecting mean leakage in the noise we provide, at our standard canvas size of $832\times1216$px.

\section{VAE Scale-and-Shift}
We disclose the scale-and-shift of our anime dataset, in hopes of encouraging a scale-and-shift of training practices. Whilst this technique was not used on NAIv3, it is a practice we would like to raise awareness of for future model training.

\subsection{Background: Current Latent Scaling Practices}
Conventionally, latent diffusion setups adapt from the VAE's latent distribution to the diffusion model's distribution like so:
\begin{itemize}
    \item A scale factor of $0.18215$ (SD1) or $0.13025$ (SDXL) is used
    \item Latents encoded by the VAE would be multiplied by this scale factor before being (noised and) given to the diffusion model.
    \item Denoised latents output by the diffusion model would be divided by this scale factor before being given to the VAE decoder.
\end{itemize}
This practice was established in \cite{rombach2022highresolutionimagesynthesislatent}, where the value comes from the (reciprocal of the) standard deviation of a single batch of images. The intention is to allow the diffusion model to train on data which has unit variance, by scaling down the comparatively high-variance VAE output. This is all the more important when training using the EDM formulation\cite{karras2022elucidatingdesignspacediffusionbased}, which is parameterized on $\sigma_{data}$ (the standard deviation of the dataset), which in latent diffusion is typically configured to 1. Scaling our data to have unit variance helps to satisfy this property we claim of our dataset.

\subsection{Proposal: Something Entirely More Gaussian}
We suggest a few changes to this practice:
\begin{itemize}
    \item Apply scale \textbf{and} shift; seek to center our data on its mean.
    \begin{description}
     \item This gives our data a mean of 0 and unit variance, making it a standard Gaussian. Mean-centering the data may help the model to work without biases in its conv\_in and conv\_out.
    \end{description}
    \item Apply a \textbf{per-channel} scale-and-shift
    \begin{description}
     \item This makes \textit{each channel} a standard Gaussian. Benefits of standardizing neural network inputs in this way are discussed in \cite{nnstandardization}.
     \item Comes with a potential downside of decorrelating the channels, which may make it harder to identify signals such as "all channels high = white". So there is a question of which benefit is preferable to have.
    \end{description}
\end{itemize}

This is not a new idea, but it is underutilized in the SDXL community. Per-channel scale-and-shift has been applied to the latent distribution of the original stable-diffusion\cite{rombach2022highresolutionimagesynthesislatent} VAE finetunes, for the training of diffusion decoders\cite{betker2023dalle3} and image-generative diffusion models\cite{Karras_2024_CVPR}. Per-channel scale-and-shift is a prevalent idea in pixel-space training, with common dataloader examples normalizing using ImageNet statistics\cite{normalizationinthemnistexample}.

To advance SDXL training practices, we share the scale-and-shift of our anime dataset (\cref{fig:vae_scale_shift}), computed over 78224 images using a Welford average\cite{doi:10.1080/00401706.1962.10490022}. This online average enables us to accumulate an average over multiple batches without precision loss or memory growth. Though it is not an entire-dataset average, it is satisfactorily converged.

\begin{figure}[H]
    \centering
    \begin{tabular}{||c||c c c c||} 
        \hline
        Variable & 0 & 1 & 2 & 3 \\ [0.5ex] 
        \hline\hline
        $\mu$ & 4.8119 & 0.1607 & 1.3538 & -1.7753 \\ 
        $\sigma$ & 9.9181 & 6.2753 & 7.5978 & 5.9956 \\ [1ex] 
        \hline
    \end{tabular}
    \caption{Scale-and-shift of latent anime illustrations encoded by SDXL VAE}
    \label{fig:vae_scale_shift}
\end{figure}

Notice how if we average $\sigma$ over all channels, we get $7.4467$, whose reciprocal is $0.1343$ — nearly the $0.13025$ with which SDXL is typically configured. Notice also how the channels are not mean-centered, and their variances… vary.

\lstset{
    frameround=fttt,
    language=Python,
    numbers=left,
    breaklines=true,
    keywordstyle=\color{blue}\bfseries, 
    basicstyle=\ttfamily\color{red},
    numberstyle=\color{black}
    }
\lstMakeShortInline[columns=fixed]|

To adapt latents encoded by the SDXL VAE into standard Gaussians: one would subtract these means ($\mu$) then divide by these stds ($\sigma$), \lstinline[columns=fixed]{encoded.sub_(means).div_(stds)}. A convenient implementation of this is available in \lstinline[columns=fixed]{torchvision.transforms.v2.functional.normalize}.

To adapt the diffusion model's standard Gaussian output to the VAE decoder distribution: one would multiply these stds ($\sigma$), then add these means ($\mu$), \lstinline[columns=fixed]{encoded.mul_(stds).add_(means)}.

The hope is that after applying this scale-and-shift: one should observe that (on average) per-channel standard deviations \lstinline[columns=fixed]{x.std((-2, -1))} are closer to 1 and that per-channel means \lstinline[columns=fixed]{x.mean((-2, -1))} are closer to 0 than before (note: we assume channels-first image data, as is conventional for convolutional models).

These statistics are dataset-dependent; illustration datasets will exhibit more variance than natural images, and if white backgrounds are well-represented then the mean of the distribution will be elevated. It is worth measuring statistics for whichever dataset is to be used, but the statistics we share should be applicable to anime illustration datasets.

\end{document}